# How Well Does AI Peer Review Work? A Benchmark Built from 100 Planted Errors

Paul Litvak
Robyn Dawes Institute
pmlitvak@gmail.com

## ABSTRACT

One hundred known errors were planted into 10 open-access psychology papers, and the modified papers were then run through frontier models and two commercial AI review tools. The best single system caught 71 of the 100 errors, while the worst caught 30. Pooling every system's output caught 93. Models are only partly correlated in the errors they find, which makes ensembling a big lever for finding issues in papers. Seven errors could not be caught by any system, and all seven were omissions: information deleted from a paper rather than mistakes inserted into it. Refine.ink contributes more unique catches than any other single system, though it is expensive on a per-error basis. False positives were not measured, and how this error distribution compares to the distribution of errors in real papers remains unknown. The papers, errors, model outputs, and full experiment log are public, so that others can build on this toward a comprehensive evaluation benchmark across disciplines.

## 1 Introduction

There has been a proliferation of AI tools that purport to do peer review. This is coming at the exact moment when the number of papers is also starting to grow due to AI assistance. All this new knowledge is about to be created, but humans lack the collective attentional capacity to review it all. We are on the precipice of a revolution in collective knowledge infrastructure, where the next step is going to surpass the span of the human mind itself. But we aren't yet able to fully turn ourselves over to it. How can we know that what it produces is right, across a given field of knowledge, as the frontier of AI continues to move (and is extremely jagged)? This problem is most urgent in the sciences. Most scientists following these tools closely would agree that AI peer review, the ability of AI to evaluate a given piece of scientific research, is beginning to get quite good.[1] But how good? And how do we measure when it's time to hand over the reins?

A medical school professor, in conversation on this topic, made the point that the new science isn't going to be legible. The best human minds in medicine have an increasingly narrow scope of in-depth knowledge of a specific subject area. Science will soon outpace the ability of any individual human mind, and the time it has available, to comprehend it. We will likely need a shared, continuously updated record of what is actually known in a field (a living context layer for science) if we as humans have any hope of keeping up. As an example, the concept of diabetes we have right now is merely a shorthand[2] for a much more complex phenomenon. It might be better represented by a 90-dimensional vector spanning dozens of biomarkers instead of a single threshold like A1C.

---

[1]The recent announcement of a partnership between the AEA, the Econometric Society, and Refine.ink to add the latter's assessments into the journal submission workflow is a portent of where things are going. See https://www.refine.ink/blog/economics-journal-partnerships.

[2]David Chapman has been writing about the nebulosity inherent in concepts like HIV and AIDS. These concepts are affordances for action, but they don't have some kind of ontological groundedness the way most people naively assume.

Correspondingly, the standard of care would stop being a threshold rule a clinician can hold in their head and become a function computed over that whole representation.

Leaning on AI for any of that requires knowing what these systems actually catch. This paper describes a benchmark built to measure that, and what it says about where current systems fail.

## 2 Creating a Benchmark

This work started as an attempt, brainstormed with Claude, to develop good evaluations of peer review software. Most systems work by ensembling, having the LLMs judge each other's output. That, as the results below show, is a good strategy, since different LLMs are surprisingly uncorrelated as far as the errors they can catch. Nonetheless, this isn't satisfying since it leaves open whether even pooled models miss some categories of errors. Using human peer review as the gold standard, as any professor will tell you, is also unsatisfying, since human peer review varies widely in quality. Of Claude's suggestions, one stood out: inserting errors into papers and seeing whether a given AI system can spot them. Although this doesn't measure false positives (identifying errors that aren't really errors), it at least provides a grounded baseline for measuring true positives and false negatives.

Given the viability of this approach, 10 existing open-access psychology papers[3] were modified[4], with errors inserted both by deleting key clauses and by altering others. A taxonomy of errors in experimental psychology papers was built first, drawn from existing taxonomies like the INSPECT-SR checklist [3]. The result was a 62-category taxonomy of types of errors, ranging from statistical errors to construct validity issues, causal inference, and generalizability. From that taxonomy, 100 errors were created, 10 per paper, and inserted.

The first iteration was a bust. The errors were too basic and the benchmark was immediately saturated when the papers were run through foundation models. So it took a few iterations to get this right. Getting high-quality errors required deleting text from papers, not just inserting mistakes. For example, one paper had a key statistical transformation that meta-analysis requires removed. In another, a sentence explaining how the key dependent measure was calculated was deleted, making it difficult to evaluate the validity of the study's claims. So the errors in some cases became quite subtle. A careful reviewer, human or AI, should notice that there's some important missing information.

## 3 Evaluation Results

The error-filled papers were then run through various versions of Claude, ChatGPT, and Gemini[5], as well as Reviewer3 and Refine.ink, two of the more well-known paid AI paper evaluation systems. An LLM-as-judge examined each of the errors each system spotted and checked whether they corresponded precisely to any of the planted errors[6]. For the foundation models, the papers were also rerun three times to check test-retest reliability, and whether running the same model multiple times could spot additional errors. It didn't, although temperature settings could have been varied and this looked at more deeply. Table 1 gives the results.

---

[3]Part of this project was also to experiment with novel ways of doing scientific research using current AI systems: using AI to help brainstorm approaches to a problem, and then relying on AI to conduct background research, write code, and evaluate outputs. The benchmark work described here was carried out by the author in collaboration with Claude.

[4]Starting with the field the author knows best. The papers came from OSF, selected for a diverse set with the most permissive open-access license and available in .docx, which made modification easier and sidestepped annoying PDF parsing issues.

[5]All runs used each provider's public API. Several of these model identifiers are dateless aliases that providers update in place, so run dates are listed too. Claude Sonnet 4.6 (claude-sonnet-4-6) and Claude Opus 4.6 (claude-opus-4-6), each with and without extended thinking, and GPT-5.4 (gpt-5.4), with and without high reasoning effort, were run March 19-20, 2026. Gemini 3.1 Pro (gemini-3.1-pro-preview) and Gemini 3 Flash (gemini-3-flash-preview) were run March 21, 2026. GPT-5.5 (gpt-5.5, with and without high reasoning) and Claude Opus 4.8 (claude-opus-4-8, with and without extended thinking) were run June 25, 2026. The commercial tools expose no version identifiers, so the run date is the only indicator of their version: Refine.ink was run June 24, 2026, and Reviewer3 (journal mode) on June 25, 2026. Reviews were scored with the pipeline as of each run; all numbers here reproduce from the stored outputs in the public repo.

[6]This also required a few iterations. The AI agent tried to use overly simplistic methods to assess error correspondence. A few cases were spot-checked, though more time could have been devoted to evaluating this.

**Table 1: The leaderboard. Cost/error is the price of the ten-paper run divided by errors caught.**

| System | Caught /100 | Findings | Cost/paper | Cost/error |
|---|---|---|---|---|
| GPT-5.5 (high reasoning) | 71 | 512 | $0.30 | $0.042 |
| GPT-5.5 (no reasoning) | 66 | 704 | $0.30 | $0.045 |
| GPT-5.4 (no reasoning) | 60 | 421 | $0.13 | $0.022 |
| Claude Sonnet 4.6 | 60 | 356 | $0.18 | $0.030 |
| Claude Opus 4.6 | 59 | 310 | $0.25 | $0.042 |
| Refine.ink | 57 | 305 | ~$50 | ~$8.77 |
| GPT-5.4 (high reasoning) | 56 | 278 | $0.32 | $0.057 |
| Claude Opus 4.8 (thinking) | 52 | 227 | $0.40 | $0.077 |
| Gemini 3.1 Pro | 42 | 203 | $0.03 | $0.007 |
| Gemini 3 Flash | 36 | 200 | $0.004 | $0.001 |
| Reviewer3 (journal mode) | 30 | 125 | subscription | n/a |

The best individual model (GPT-5.5) only caught 71 of the 100 errors, while the worst, Reviewer3, caught 30. Since there have been a number of model upgrades (and who knows how much Refine and Reviewer3 have upgraded in the months since these runs), the specific eval hierarchy matters less than the overall pattern of results. The sample size of errors is also small enough that a gap of a few errors isn't meaningful, so the ordering of systems that scored close together shouldn't be read into. Note also that the findings column indicates the number of review comments overall. Since the baseline papers themselves have issues that were not assessed, whether those comments are correct is unknown. It's worth noting the trend, however, that more successful systems surfaced more issues in general.

One of the most notable findings is that ensembling improves error catching dramatically. Pooling errors across the foundation models catches 91 of the 100 errors, and pooling every system gets to 93 (Figure 1). Refine.ink adds the largest contribution to the ensemble over and above the best single model.

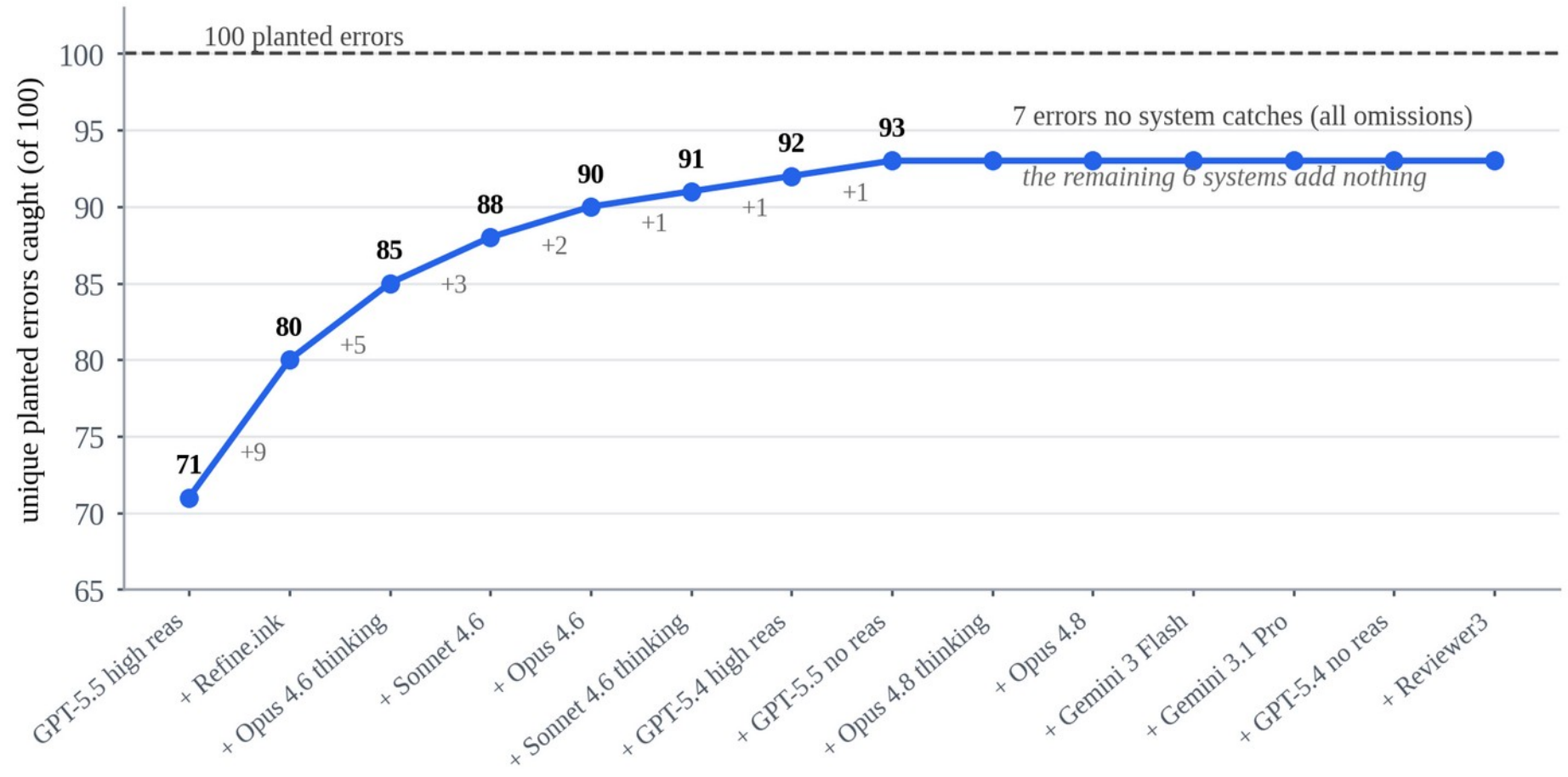

Figure 1: Ensemble coverage saturates fast. Systems are added greedily; each step adds the system that catches the most errors the ensemble still misses.

Refine.ink (and Reviewer3 to a lesser extent) do add unique error checking capability above and beyond what the foundation models provide. Refine is very expensive on a per-error basis ($8.77 for each of the errors caught, against about four cents for the frontier models), but does definitely seem to add unique value. It's a leader in

detecting issues around methodological design, statistical errors, and internal consistency, but among the weakest on causal inference, reporting completeness, and generalizability. Again, this may have improved in recent months. It does suggest that with better evals and added compute to address these gaps, Refine.ink likely would be better than the foundation models.

All the systems struggle with the omission-based errors. There was a small set of 7 errors that no system was able to detect. In one paper, for example, the open science disclosures around pre-registration were dropped, along with the statement that all measures, manipulations, and exclusions were included, making it possible that the paper engaged in p-hacking. No system proactively called that out. A careful reviewer would certainly ask whether there are any unreported measures, or ask the authors to disclose it. This relates to AbsenceBench [1], which tested whether models can identify which pieces of a document were deliberately removed, given access to both the original and edited documents. Models are bad at this, and that's with both versions in hand. The systems evaluated here only saw the edited paper.

## 4 Caveats, Limitations, Next Steps

One caveat is that verification of the results could have been more thorough. The errors were spot-checked, Claude was asked to explain issues that were unclear, and the set was iterated on and rechecked several times. All 10 papers were not, however, read in painstaking detail with all 100 errors checked by hand. The goal was to demonstrate the method in sufficient detail that a larger team could build upon it. Verification continues to be the bottleneck, both for this work specifically and in general. Even if a few of the inserted errors are wrong, directionally the results are likely to be correct, though some unease about that remains. To that end, the repo is public for anyone to download and examine. It includes a detailed experiment log that explains every iteration on the benchmark and the evaluation runs, the original and modified papers, the AI-generated review comments, and a writeup iterated on with Claude. Publishing this benchmark kills its utility, since it should be straightforward to find 10 new papers and do the same exercise. There's more value in demonstrating the method than in keeping this specific benchmark secret.

It's important to note that this was only for empirical psychology papers, so it's far from obvious this would generalize to other academic subfields. Chenhao Tan's research group at UChicago recently used the same methodology of injecting errors into papers across eight arXiv subject classes and found very similar results, with their best configuration recovering 71.6% of injected errors [2]. The hope is to find collaborators interested in expanding this eval for more areas of science, creating high-quality, fully vetted errors to insert, and building a public benchmark. If we are going to rely on these systems to screen papers, public evaluation of them is essential.

Another limitation is that nothing here speaks to false positives, errors that the systems flag that aren't real errors. False positives are known to be an issue. Running many papers through foundation models surfaces illogical or fake errors that have to be pushed back on. Recall is now measurable; precision is not. Measuring precision requires something closer to a human gold standard, with experts manually verifying the validity of each flagged issue. That information would be very valuable to collect.

Finally, the most important limitation is that how well these errors generalize to the kinds of errors found naturally in empirical papers is unknown. This mismatch could come in two flavors. First, the errors created here may not be like errors really found in papers. Second, the error distribution in actual empirical papers is itself unknown. Errors here were sampled uniformly across category, but it may be that certain types of errors are more common than others. It's possible that the kinds of errors current AI systems are bad at catching are the most important and common errors, while the obvious ones are less common or less important. Ideally the benchmark results would be weighted by the frequency of errors found in a random sample of empirical papers. This is challenging and requires some reliable expert human coding to get. One obvious approach is to use the best peer review system (or the union of all of them) to generate errors, categorize them, and use the inverse of the false negative rate per error category as a multiplier. For example, if only 10% of some error type is caught on average, assume there are 10x more errors of that type and use that to estimate the empirical distribution. This breaks down when there are error types no system can catch, so unfortunately it probably won't work just yet.

## 5 Conclusion

AI peer review is here whether we like it or not. AI-generated science is also here and will increasingly be a larger and larger portion of new scientific knowledge. That knowledge will quickly outstrip the ability of humans to evaluate it. Amidst that new knowledge (especially at first), there are going to be a considerable number of mistakes. We need to start thinking very carefully about how to surmount the evaluation and human expert attention bottleneck. Training the next generation of scientists with the requisite domain expertise, critical thinking, and AI-native thinking is essential if we have any hope of avoiding being drowned in all the new information, or of preventing it from being riddled with errors and mistakes.

Finally, it's important to consider that all of these AI peer review systems are missing at present a real living context layer that allows them to evaluate a given paper in light of everything else that's known about the topic.[7] To some degree that information is encoded in the weights of foundation models, but these models are trained on all papers and don't yet have the ability to weight the balance of evidence the way an expert would. Even if current systems become essentially perfect at catching errors, a public record of the current best understanding of every scientific hypothesis is still required to evaluate and contextualize new findings. To really close the gap with expert human reviewers, we need to build that too.

## Data and Code Availability

The modified papers, the planted errors, all model outputs, the scoring pipeline, and the full experiment log are at https://github.com/Dawes-Institute/ai-peer-review-benchmark.

## References

[1] Harvey Yiyun Fu, Aryan Shrivastava, Jared Moore, Peter West, Chenhao Tan, and Ari Holtzman. 2025. AbsenceBench: Language Models Can't Tell What's Missing. arXiv:2506.11440.

[2] Dang Nguyen, Wanqing Hao, Yanai Elazar, and Chenhao Tan. 2026. Benchmarking Agentic Review Systems. arXiv:2606.19749.

[3] Jack Wilkinson, Calvin Heal, George A. Antoniou, et al. 2023. Protocol for the development of a tool (INSPECT-SR) to identify problematic randomised controlled trials in systematic reviews of health interventions. medRxiv 2023.09.21.23295626.

[7]The need for such a context layer, and the path toward creating it, is discussed in previous meta-science posts at paullitvak.com.